\pdfoutput=1

\documentclass[11pt]{article}

\usepackage[final]{acl}

\usepackage{times}
\usepackage{latexsym}

\usepackage[T1]{fontenc}

\usepackage[utf8]{inputenc}

\usepackage{microtype}

\usepackage{inconsolata}

\usepackage{graphicx}
\usepackage{amssymb}
\usepackage{calrsfs}
\usepackage{amsmath}
\usepackage{booktabs}
\usepackage{multicol}
\usepackage{multirow}

\DeclareMathAlphabet{\pazocal}{OMS}{zplm}{m}{n}

\newcommand{\Lb}{\pazocal{L}}
\title{Revisiting Early Detection of Sexual Predators via Turn-level Optimization}

\author{First Author \\
  Affiliation / Address line 1 \\
  Affiliation / Address line 2 \\
  Affiliation / Address line 3 \\
  \texttt{email@domain} \\\And
  Second Author \\
  Affiliation / Address line 1 \\
  Affiliation / Address line 2 \\
  Affiliation / Address line 3 \\
  \texttt{email@domain} \\}

\author{Jinmyeong An$^1$, Sangwon Ryu$^1$, Heejin Do$^1$,\\ \textbf{Yunsu Kim}$^3$,  \textbf{Jungseul Ok}$^{1,2}$, \textbf{Gary Geunbae Lee}$^{1,2}$\\
        $^1$Graduate School of Artificial Intelligence, POSTECH, Republic of Korea \\
	 $^2$Department of Computer Science and Engineering, POSTECH, Republic of Korea \\
    $^3$aiXplain Inc., Los Gatos, CA, USA \\
    \texttt{\small \{jinmyeong, ryusangwon, heejindo, jungseul.ok, gblee\}@postech.ac.kr}, \texttt{\small yunsu.kim@aixplain.com}
    }

\begin{document}
\maketitle

\begin{abstract}

Online grooming is a severe social threat where sexual predators gradually entrap child victims with subtle and gradual manipulation. Therefore, timely intervention for online grooming is critical for proactive protection. However, previous methods fail to determine the optimal intervention points (i.e., jump to conclusions) as they rely on chat-level risk labels by causing weak supervision of risky utterances. 
For timely detection, we propose speed control reinforcement learning (SCoRL)\footnote{The code and supplementary materials are available at \url{https://github.com/jinmyeongAN/SCoRL}}, incorporating a practical strategy derived from luring communication theory (LCT). To capture the predator's turn-level entrapment, we use a turn-level risk label based on the LCT. Then, we design a novel speed control reward function that balances the trade-off between speed and accuracy based on turn-level risk label; thus, SCoRL can identify the optimal intervention moment. In addition, we introduce a turn-level metric for precise evaluation, identifying limitations in previously used chat-level metrics. Experimental results show that SCoRL effectively preempted online grooming, offering a more proactive and timely solution. Further analysis reveals that our method enhances performance while intuitively identifying optimal early intervention points.
\end{abstract}

\section{Introduction}

Online grooming is a manipulative tactic where sexual predators establish emotional connections with minors over the internet to exploit them for harmful purposes \cite{wachs}. 
Such exploitation is particularly insidious, as it often leads victims to meet their abusers offline voluntarily, resulting in severe risks \cite{olson2007entrapping, cano2014detecting}. 
Therefore, early and timely intervention within the dialogue is essential for sexual predator detection (SPD) \cite{villatoro2012two}.

While conventional SPD approaches have primarily focused on post-analysis to identify completed grooming cases for legal prosecution \cite{bours2019detection,inches2012overview,ebrahimi2016detecting}, recent efforts have shifted towards early sexual predator detection (eSPD) \cite{PANC}, which aims to detect risks in real-time and intervene before abuse escalates. \citet{escalante2015early, monroy2018early} framed eSPD as an early text classification task, and \citet{PANC} employed a rule-based classification that continuously evaluates risk using segmented chat (dialogue) data as training data.

\begin{figure}
\centering
\includegraphics[width=\linewidth]{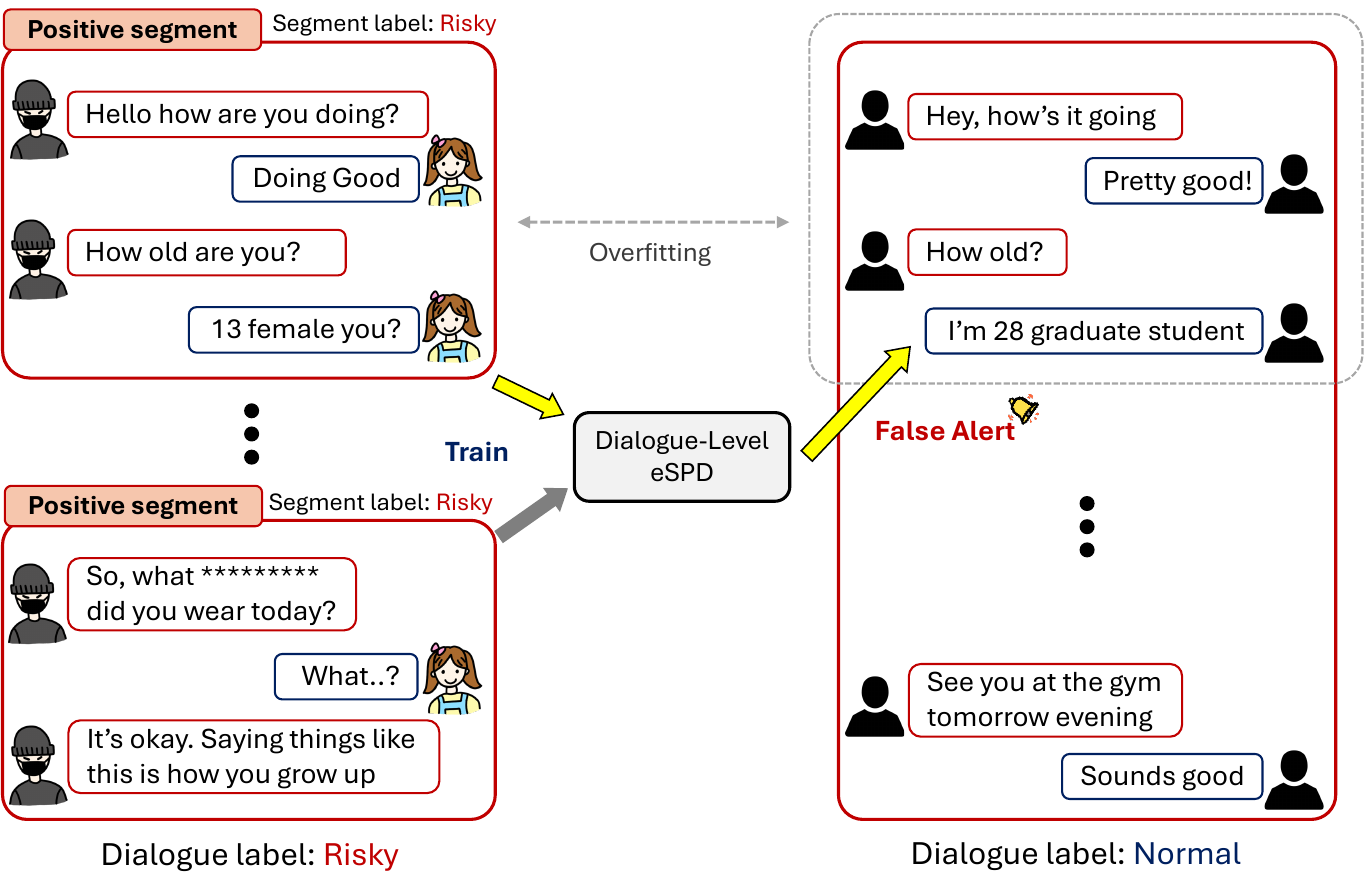}
\caption{Yellow arrows highlight the weakness of dialogue-level (chat-level) supervision in accurately identifying risky utterances. Previous approaches segment long conversations and assign the same dialogue-level label to each segment. However, risky utterances are often sparse within these segments, leading to mislabeling and increased false positives during normal conversations. }
\label{fig: fig1}
\end{figure}

Although various eSPD methods have been proposed, most rely solely on chat-level labels, which provide weak supervision and fail to identify specific risky utterances. This leads to significant detection errors, such as false alarms in normal conversations (also referred to as dialogues or chats), as illustrated in Figure \ref{fig: fig1}. The reliance on chat-level labels results in overfitting, causing the model to misclassify normal dialogues as high-risk due to the lack of granular information.
The use of chat-level labels also hinders model optimization, preventing it from identifying the precise moment where early detection should occur. 
To address the instability caused by weak supervision, some previous works have employed count-based methods \cite{PANC}, which use a sliding window over the messages of a chat and classify them. However, such rule-based approaches are limited in their ability to create an optimal early detection model, as they lack the flexibility and sophistication to adjust dynamically to real-time risks.

In this work, we revisit the task of eSPD by incorporating turn-level labels, which provide finer granularity and more reliable supervision. We utilize a turn-level labeled dataset \cite{mcghee2011learning, cano2014detecting} that is annotated based on the Luring Communication Theory (LCT) \cite{olson2007entrapping}, capturing the strategies predators use at each stage of the conversation.
In addition, we propose a novel Speed-Controlled Reinforcement Learning (SCoRL) method designed to optimize the identification of the optimal early detection point. Our SCoRL framework leverages a speed-control reward function, balancing detection timeliness and accuracy.
Moreover, we introduce a new benchmark, Turn-Level eSPD, to evaluate eSPD models more effectively. Unlike traditional metrics, our proposed metric accounts for both turn-level and chat-level risk factors, ensuring a more precise evaluation.
In experiments on the PANC dataset \cite{PANC}, our SCoRL method significantly outperforms existing approaches in terms of eSPD performance.
Through comprehensive visualizations and case studies on real online grooming conversations, we show that our model not only improves performance quantitatively but also intuitively identifies the optimal points for early intervention.

Contributions:
(1) We propose an effective eSPD method using reinforcement learning with a novel speed-controlled reward, leveraging turn-level annotations of the predator's entrapment strategy for more granular supervision.
(2) We introduce a new benchmark, Turn-Level eSPD, which enables more accurate evaluations by addressing the limitations of previous metrics that overlook false positives.
(3) We empirically demonstrate the superiority of our approach through extensive experiments, highlighting both the quantitative and qualitative benefits.

\section{Related Work}

\subsection{Sexual Predator Detection}
Online grooming conversations are designed to build trust and persuade the victim to engage in dangerous offline interactions \cite{olson2007entrapping}. Detecting such behaviors early is crucial.
Existing Sexual Predator Detection (SPD) methodologies primarily focus on retrospective identification of predators through chat segment classification, either at the author level \cite{villatoro2012two, cardei2017detecting} or using binary classification over chat segments or entire conversations \cite{ebrahimi2016detecting, bours2019detection}. Much of this work has been evaluated using the PAN shared task dataset \cite{inches2012overview}.

However, most prior research focuses on detecting grooming only after the conversation is complete. Early Sexual Predator Detection (eSPD) has received comparably less attention. Notably, \citet{escalante2015early} introduced early text classification for SPD, followed by \citet{monroy2018early}, who improved segment classification using Multi-Resolution Representations (MulR).
A more formal eSPD approach was proposed by \citet{PANC}, using chat-level labels and a rule-based classifier. However, due to the weak supervision of chat-level labels, it is not possible to know which specific utterances are dangerous.
Our work addresses this gap by leveraging turn-level strategy labels to accurately capture risk utterances and optimize early detection.

\subsection{Early Risk Detection}
Early Risk Detection (ERD) focuses on identifying harmful or dangerous behaviors at their earliest stages \cite{losada2018overview}. Several studies have explored ERD across various domains, such as detecting early signs of depression \cite{depressionEarly}, self-harm \cite{ragheb2019attentive}, and anorexia \cite{paul2018early}.
However, many ERD techniques rely heavily on rule-based systems \cite{ma2016detecting}.
More recent efforts, such as those by \citet{zhou2019early} and \citet{zeng2023early}, have addressed these limitations by employing Deep Q-Learning and Neural Hawkes Processes to automatically identify detection points. 
However, a critical distinction in our work is the nature of eSPD systems, which should not classify a conversation as non-grooming as long as messages are still ongoing or expected. In contrast, existing methods \cite{zhou2019early, loyola2018learning} may prematurely decide it is safe to stop monitoring a post or chat once they deem the risk has passed. 
To address these gaps, our work introduces a novel methodology for eSPD that integrates ERD over time. This enables our model to balance detection speed with accuracy.

\section{Problem Formulation: Turn-Level eSPD}

In this section, we propose a new benchmark, turn-level eSPD. We define the turn-level risk label by incorporating the turn-level strategy derived from LCT \cite{olson2007entrapping}. We give a formal definition of the task and an evaluation setup for the turn-level eSPD. 

\subsection{Turn-Level Risk Label}
We annotate the turn-level risk label $y_t^{turn}$, utilizing four different turn-level strategy labels $s_t^{turn}$ annotated by ChatCoder2 (CC2) \cite{mcghee2011learning} dataset based on LCT.

From the LCT perspective, predator behavior typically evolves through distinct stages. In the early stage, known as \textbf{\textit{exchange of Personal Information} (\textit{PI})} stage, predators aim to build trust and intimacy with potential victims by mimicking ordinary conversations. As the dialogue progresses, the interaction may shift to more dangerous stages: \textbf{\textit{Grooming} (\textit{G})} and \textbf{\textit{Approach} (\textit{A})}, where predators escalate the interaction by introducing sexual topics and eventually soliciting physical contact. Additionally, a category labeled \textbf{\textit{Lines containing none of the classes} (\textit{Others})} includes innocent expressions unrelated to grooming activities.




We assign the turn-level risk label $y_t^{turn} = 0$ for a turn (utterance) $u_t$ if the turn-level strategy $s_t^{turn}$ corresponds to non-risky stages \{\textit{PI}, \textit{Others}\}. Following \cite{mcghee2011learning}, turns with \textit{Others} label are classified as innocent turns unrelated to grooming activities. Additionally, the \textit{PI} stage reflects conversational patterns that can occur in regular dialogues.
In contrast, we assign the turn-level risk label $y_t^{turn} = 1$ for a turn $u_t$ if the turn-level strategy $s_t^{turn}$ belongs to the dangerous stages \{\textit{G}, \textit{A}\}. \textit{G} and \textit{A} are deemed risky because they initiate and escalate \textit{the entrapment cycle} \cite{mcghee2011learning}, gradually manipulating the victim. This classification effectively distinguishes grooming conversations from normal interactions. By focusing on \textit{G} and \textit{A} as clear risk indicators, we aim to reduce false positives in our analysis.


\subsection{Task Definition}

We denote the dataset as $D = \{(C^{(1)}, S^{(1)}, y^{(1)}), \ldots, (C^{(L)}, S^{(L)}, y^{(L)})\}$, where $L$ represents the number of chats. Each instance in the $D$ comprises a chat $C$, a sequence of turn-level risk label $S$, and chat-level risk label $y \in \{0, 1\}$. $y=1$ indicates a grooming (positive) chat, while $y=0$ signifies a normal (negative) chat.
Each chat $C$ is represented as a sequence of turns $\{u_1, \ldots, u_{n_c}\}$ and $S$ is described as $\{y^{turn}_1, \ldots, y^{turn}_{n_c}\}$, where $n_c$ is the total number of turns in each $C$. In normal chat $y=0$, all $y_t^{turn}$ are annotated to 0 as a non-risky label.

In turn-level eSPD task, the system processes turns sequentially ($u_1,...,u_{n_c}$) and determines whether an early detection should be raised at each turn $u_t$. If a risky turn ($y_t^{turn}=1$) is identified, the system triggers early detection and immediately halts further processing. Otherwise, it continues until the final turn $u_{n_c}$.
If no risky turn is identified by the end of the conversation, the system classifies the chat as normal.
The goal of turn-level eSPD is twofold: (1) to detect grooming behavior as early as possible by raising an alert at the risky turn in grooming chats ($y=1$), and (2) to avoid false alarms by not triggering detection in normal chats ($y=0$) until the conversation ends.

\paragraph{Unbalanced Label Ratio}
The PANC dataset \cite{PANC} presents a challenge due to its highly imbalanced label distribution—11,802 normal conversations versus just 69 grooming conversations annotated with turn-level strategy. This imbalance leads to an increase in false positives, which negatively impacts precision.
To address this, we additionally created a more balanced test set by randomly sampling normal conversations to achieve a 10:1 ratio between normal and grooming conversations, following the setup of previous work \cite{PANC}. The details of data distributions are in Table \ref{tab: data_statistics}.

\subsection{Evaluation Metrics for Turn-Level eSPD}
In turn-level eSPD, two key objectives should be balanced: early detection and accurate detection. These goals often conflict: issuing early warnings causes lower accuracy due to limited information, whereas delaying warnings increases accuracy but diminishes earliness.

\paragraph{Turn-Level Accuracy}
Accuracy has been the primary metric in related works on sexual predator detection (SPD) \cite{bours2019detection}. However, in turn-level eSPD, the task is considered successful when early detection occurs at turn $u_t$ with $y_t^{turn}=1$ in grooming chat ($y=1$) or early detection does not happen until the end of the turn in normal chat ($y=0$).
We evaluate accuracy using the traditional metrics of precision, recall, and F1-score.

\paragraph{Earliness}
Earliness in turn-level eSPD is measured based on the number of turns exchanged before early detection is raised, also known as \textit{warning latency}. To balance accuracy and earliness, we employ a latency-weighted F1 metric \cite{sadeque2018measuring, PANC}, which penalizes delayed warnings.
The penalty for each warning latency $l \geq 1$ is calculated as follows:
\begin{equation}\label{eq1}
\textnormal{penalty}(l) = -1 + \frac{2}{(1 + \exp(-p  (l - 1)))}
\end{equation}

where $p$ controls how rapidly the penalty increases as $l$ rises. Early detection is raised immediately after the first turn incurs no penalty, while penalties increase as $l$ grows, approaching a maximum value of 1.
For each chat, we calculate the speed of successful early detection as follows:

\begin{small}
\begin{equation}
{speed} = 1 - \textnormal{median}\{\textnormal{penalty}(l) | l \in \textnormal{latencies} \}
\end{equation}
\end{small}
Finally, the latency-weighted F1 is computed as:
\begin{equation}\label{eq3}
F_{latency} = F1 \cdot speed
\end{equation}
A turn-level eSPD system is considered superior if it achieves a higher $F_{\text{latency}}$ score for a given dataset. As suggested by \citet{PANC}, speed is computed only for grooming chats that are correctly classified as such.

\begin{figure*}
\centering
\includegraphics[width=\linewidth]{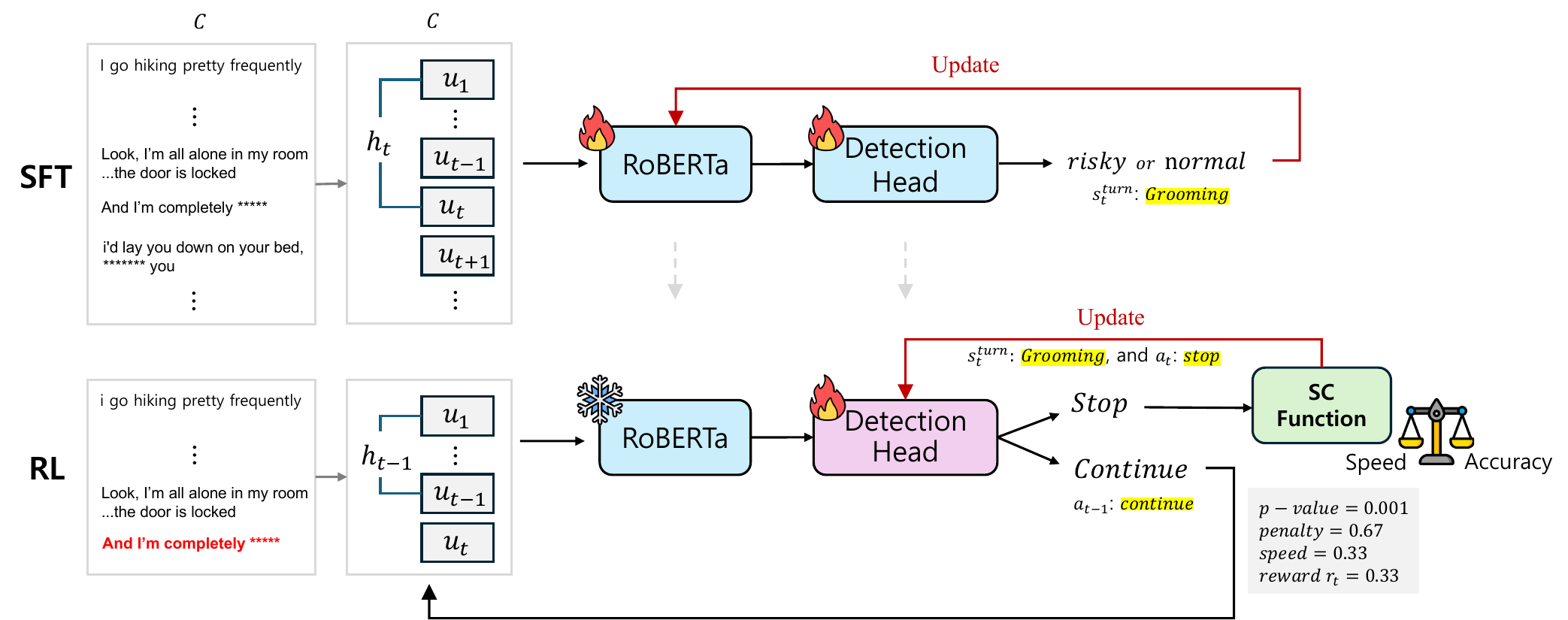}
\caption{The training overview of SCoRL. In conversation $C$, the dialogue history $h_t$ up to the current time step $t$ is input sequentially. At each step, the Supervised Fine-Tuning (SFT) model is trained using the turn-level risk label $y_t^{turn}$ for all turns.
Unlike SFT, the Reinforcement Learning (RL) process updates only the gradient of the detection head.
When $a_t = 1$, early detection is triggered, and subsequent turns are excluded from training.
A $p$ value is calculated for each conversation, and the model is updated based on the speed-control(SC) reward mechanism.}
\label{fig: main}
\end{figure*}

\section{Methodology: SCoRL}

\subsection{MDP Environment}
We model the turn-level eSPD task as a Markov Decision Process (MDP). At each turn $u_t$, based on the observed dialogue history $h_t = \{u_1, \ldots, u_t\}$, the early detection system selects an action $a_t \in \{0, 1\}$, where $a_t=0$ represents the action to "continue" to progress the conversation, and $a_t=1$ represents the action to "stop" the conversation and raise an early detection.
The system continues to read the dialogue until the system selects the "stop" action or until the system reaches the last turn $u_{n_c}$ in the dialogue.
The episode is ended immediately at $T$ when the system stops to read the dialogue, with no further turns processed.
The system aims to learn a policy $\pi$ that maximizes the expected cumulative reward across the dialogue. Formally, the optimal policy $\pi^*$ is defined as:
\begin{eqnarray}
\pi^{*} = arg\,  \mathrm{max}_{\pi \in \Pi}	\left[ \sum_{t=1}^{T} r(h_t, a_t) \right],
\end{eqnarray}
where $r(h_t, a_t)$ is the reward function that measures the appropriateness of the action $a_t$ given the dialogue history $h_t$.

\subsection{Model Training}
As illustrated in Figure \ref{fig: main}, our model aims to decide at each turn $u_t$ whether to stop the conversation to protect potential victims from online grooming. The model's decision process can be enhanced using a combination of supervised fine-tuning (SFT) and reinforcement learning (RL).
\paragraph{Supervised Fine-Tuning (SFT):}
To initialize the detection policy, we leverage a pre-trained language model (e.g., RoBERTa) with a detection head, designed to predict the turn-level risk label $y_t^{turn}$ for early detection in online grooming dialogues. We fine-tune the model using supervised learning based on a turn-level eSPD dataset $D$.
For each turn $u_t$, our SFT minimizes the cross-entropy loss between the action $a_t$ and the turn-level risk label $y_t^{turn}$, given the dialogue history $h_{t}$. The action $a_t$ should be 1 when the corresponding turn $u_t$ is risky $y_t^{turn}=1$. The SFT objective is as follows:
\begin{eqnarray}
a_t = \mathrm{\textbf{EDM}}(u_1, \ldots, u_t) \\ 
\Lb = - \frac{1}{\lvert D \rvert} \sum_{C \in D} \frac{1}{n_c} \sum_{t=1}^{n_c} a_t\, \mathrm{log} y_t
\end{eqnarray}
where $\textbf{EDM}$ is the early detection model that predicts whether to stop at turn $u_t$ or not.
While the SFT process does not directly identify the optimal turn $u_{t^*}$ for early detection, such initialization accelerates the policy convergence during reinforcement learning, as the model learns to stop the conversation when a dangerous turn is detected.

\paragraph{Speed-Control Reward} 
We introduce a speed-control reward designed to favor early and accurate detection while penalizing overly hasty decisions.
\begin{equation}
r_t =
\begin{cases} 
      speed, & \text{if } a_t = 1 \text{ and } y_t^{turn} = 1 \\
      -speed, & \text{if } a_t = 1 \text{ and } y_t^{turn} = 0 \\
      0, & \text{if } a_{t<n_c} = 0 \\
      +1, & \text{if } a_{n_c} = 0 \text{ and } y = 0 \\
      -1, & \text{if } a_{n_c} = 0 \text{ and } y = 1
\end{cases}\label{eq: speed_reward}
\end{equation}
Our reward scheme employs a delayed reward setting and encourages the model to detect risks as early as possible while avoiding hasty early detection by leveraging the \textit{speed}. However, in the case of the reward in the last turn $u_{n_c}$, we empirically find out that insufficient reward for waiting until the end of the normal chat $y=0$ often results in premature "stop", potentially causing false positives. To address this, we set the reward for appropriate waiting behavior to +1, the maximum value of \textit{speed}. Conversely, in risky conversations, failing to stop the conversation before the end poses a serious risk, as it may allow the predator to achieve their harmful objective. To prevent this, we assign -1, the minimum speed value, discouraging delayed detection in risky cases.

\begin{table}[t]
\scalebox{0.6}{
\begin{tabular}{l|c|c|c|c|c}
\toprule
 & All & \# Negative & \# Positive & Len(Negative) & Len(Positive) \\
\midrule
Train & 596 & 519 & 77 & 121 ($\pm$14) & 1430 ($\pm$2202) \\
Valid & 61 & 52 & 9 & 118 ($\pm$11) & 1340 ($\pm$802) \\
Test & 11802 & 11733 & 69 & 36 ($\pm$26) & 2050 ($\pm$3139) \\
Test w/ \textit{ds} & 759 & 690 & 69 & 36 ($\pm$25) & 2050 ($\pm$3139) \\
\bottomrule
\end{tabular}
}
\caption{Statistics of PANC dataset for turn-level eSPD. \textit{ds} denotes downsampled.} \label{tab: data_statistics}
\end{table}

\paragraph{Speed Control RL (SCoRL)}
We denote the EDM policy as $\pi(a_{t}|h_{t})$, which returns the probability of taking action $a_t$ given the state $h_t$. To optimize the policy, we utilize the vanilla policy gradient method \cite{NIPS1999_464d828b} rather than a value-based method since $r_{t<T}=0$ and only $r_T$ is non-zero at the end of episode $T$, formulated as follows:
\begin{equation}
    \theta \leftarrow \theta - \alpha \nabla\, \mathrm{log} \pi_{\theta}(a_t|h_t)R_t
\end{equation}
where $\theta$ denotes the policy network parameter, $\alpha$ denotes the policy network learning rate, and $R_t$ is the total reward accumulating from turn $u_t$ to the last episode turn $u_T:\, R_t=r_T$.
Through reinforcement learning using this speed-control reward structure, the system learns an optimal early detection policy. This enables the model to identify the optimal turn $u_{t^*}$ for detecting risky behavior, effectively distinguishing between grooming and regular conversations.
The model, initially trained via SFT, is further refined using RL, following the approach of \citet{plug}. However, only the gradient of the detection head is updated. This selective update process, inspired by \cite{lora}, ensures both efficient and stable training, making it suitable for real-world applications, as discussed in Appendix \ref{sec:Computational cost}.

The parameter $p$, which controls how quickly the \textit{speed} in the speed-control reward increases, is calculated individually for each conversation, as suggested by \citet{PANC}. For positive chats, following prior studies \cite{PANC}, $p$ is determined when the penalty is set to 0.5, with the latency $l$ defined as the point where 20 risky utterances have accumulated. For negative chats, as in \citet{losada2018overview}, $p$ is computed under the assumption of a 0.5 penalty, with the latency $l$ defined as the median length across all negative chats.


\begin{table*}[]
\centering
\scalebox{0.70}{
\begin{tabular}{l|c|cccc|c|cccc}
\toprule
\multirow{2}{*}{Model} & \multicolumn{5}{c}{Turn-level eSPD (Original)} & \multicolumn{5}{c}{Turn-level eSPD (Downsampled)} \\ \cmidrule(rl){2-6} \cmidrule(rl){7-11}
 & Latency F1  & Precision & Recall & F1 & Speed & Latency F1 & Precision & Recall & F1 & Speed \\
\midrule
$S_\text{BERT-large}$ \cite{PANC} & 0.034  & 0.025 & 1.000 & 0.049 & 0.693 & 0.122 & 0.096 & 1.000 & 0.176 & 0.693 \\ 
$S_\text{BERT-base}$ \cite{PANC} & 0.062  & 0.047 & 1.000 & 0.089 & 0.688 & 0.217 & 0.188 & 1.000 & 0.316 & 0.689 \\ 
$S_\text{mobileBERT}$ \cite{PANC} & 0.025  & 0.017 & 1.000 & 0.034 & 0.734 & 0.134 & 0.100 & 1.000 & 0.182 & 0.734 \\ 
\midrule
SFT & 0.002  & 0.001 & 1.000 & 0.002 & 0.940 & 0.023 & 0.012 & 1.000 & 0.024 & 0.940 \\ 
SFT (w/ turn-level) & 0.211 & 0.202 & 1.000 &  0.335 & 0.631 & 0.298 & 0.310 & 1.000 & 0.473 & 0.631 \\ 
ScoRL & 0.000  & 0.000 & 0.182 & 0.000 & 1.000 & 0.005 & 0.003 & 0.182 & 0.005 & 1.000 \\ 
ScoRL (w/ turn-level) & \textbf{0.365} &  0.475 & 0.983 & 0.641 & 0.569 & \textbf{0.509} & 0.819 & 0.983 & 0.894 & 0.569 \\ 
\bottomrule
\end{tabular}%
}\caption{Turn-level eSPD performance on the original and downsampled PANC dataset.}\label{tab: new_metric}
\end{table*}



\begin{table}[t]
\centering
\scalebox{0.64}{
\begin{tabular}{l|c|cccc}
\toprule
Model & Latency F1 & Precision & Recall & F1 & Speed \\
\midrule
$S_\text{BERT-large}$ & 0.247 & 0.217 & 1.000 & 0.358 & 0.693 \\
$S_\text{BERT-base}$ & 0.244 & 0.215 & 1.000 & 0.355 & 0.688 \\
$S_\text{mobileBERT}$ & 0.156 & 0.119 & 1.000 & 0.212 & 0.734 \\
\midrule
SFT  & 0.018 & 0.010 & 1.000 & 0.019 & 0.939 \\
SFT (w/ turn-level) & \textbf{0.489} & 0.633 & 1.000 & 0.775 & 0.631 \\
ScoRL  & 0.010 & 0.005 & 0.870 & 0.010 & 1.000 \\
ScoRL (w/ turn-level) & 0.401 & 0.548 & 0.986 & 0.705 & 0.569 \\
\bottomrule
\end{tabular}
}
\caption{eSPD performance on the PANC dataset based on previous chat-level metric.}\label{tab: previous_metric}
\end{table}

\section{Experimental Setup}

\paragraph{Dataset}
We utilize the publically available PANC dataset \cite{PANC}, designed for the eSPD task, which is based on PAN12 \cite{inches2012overview} and ChatCoder2 (CC2) \cite{mcghee2011learning, cano2014detecting}. The dataset consists of three types of corpus: (1) positive (grooming) full-length chats from CC2, (2) negative (normal) segments from PAN12, and (3) positive segments that split the positive full-length chats into parts.
Our experiments focus on training and evaluating the model using only the positive full-length chats and the negative segments. We use the turn-level strategy annotations from CC2, as labeled by \citet{mcghee2011learning}, who assigned LCT strategy labels to both predator and victim turns when their content aligned with the LCT framework. According to PAN12, negative segments include cybersex conversations between consenting adults, which serve as hard negative samples since they closely resemble online grooming chats (see Appendix \ref{sec:FP in normal}). Additionally, they involve cutting the conversations at points where the exchange exceeds 25 minutes or marks a topic shift. Therefore, we treat the negative segments as negative full-length chats for training and testing purposes.
Table \ref{tab: data_statistics} summarizes the PANC dataset statistics.

\paragraph{Baseline Models} 
We follow the two-tier method proposed by \citet{PANC}, representing the current state-of-the-art for the eSPD task.
In Tier 1, we fine-tune three variants of BERT—BERT-large, BERT-base \cite{bert}, and MobileBERT \cite{mobilebert}—using the positive and negative segments labeled at the dialogue level, same with \citet{PANC}. During evaluation, each model takes a sliding window of messages (window size = 50) and outputs a binary classification for grooming or non-grooming.
In Tier 2, the model monitors the number of positively classified windows in the last 10-window span. If the count surpasses a predefined threshold, referred to as "skepticism" (set to 5 in our setup), the conversation is classified as grooming.
We trained the models using a cross-entropy loss function with the Adam optimizer \cite{kingma2014adam}. The batch size was set to 16, with a training run for 10 epochs. The learning rates were $3 \cdot 10^{-5}$ for BERT-large, and $5 \cdot 10^{-3}$ for both BERT-base and MobileBERT.

\paragraph{SFT Model}
Our SFT model was trained using turn-level risk labels for each turn. The dataset \cite{PANC} exhibited significant class imbalance, with negative chats outnumbering positive chats with turn-level risk label by more than 100:1. To mitigate this imbalance, we sampled negative chats for training, prioritizing conversations with more than 100 turns to enrich the turn-level information (see Table \ref{tab: data_statistics} for details).
During training, the model processed up to 50 utterances of conversation history and a turn-level risk label for the current utterance. We used a RoBERTa-base model \cite{roberta} as the binary classifier, which consists of 12 transformer encoder layers, each with 768 hidden units and 12 attention heads. A detection head with dimensions [768, 768, 2] was attached to perform binary classification at the utterance level.
The model was trained with cross-entropy loss using the Adam optimizer \cite{kingma2014adam} for 10 epochs, with a batch size of 32 and a learning rate of 5e-05.

\paragraph{SCoRL Model}
The SCoRL model continues processing turns in an episode until either $a_t = 1$ or the final turn of the conversation is reached. For each turn, the current utterance and the previous 50 utterances in the conversation history are concatenated, separated by special tokens, forming the current state.
The RoBERTa-base encoder, pre-trained in the SFT step, is used with frozen parameters. Only the detection head ([768, 768, 2]) is trained during this step. The model was trained with a batch size of 32 for 50 epochs, using a learning rate of 2e-4, a discount factor of 0.99, and the Adam optimizer. We employed the CosineAnnealingWarmUpRestarts \footnote{\url{https://pytorch.org/docs/stable/generated/torch.optim.lr_scheduler.CosineAnnealingWarmRestarts.html}} scheduler with the following parameters: $T_0=10$, $T_{mult}=1$, $eta_{max}=0.1$, $T_{up}=1$, and $gamma=0.5$. 

\section{Result and Analysis}

\subsection{Overall Turn-Level eSPD Performance}
The primary goal of turn-level eSPD is to ensure both high accuracy and low latency in distinguishing between online grooming and normal conversations, thereby enabling the timely detection of risky utterances. To verify our method, we compare our model with the previous eSPD systems regarding Latency-F1, F1-score, precision, and recall (Table \ref{fig: main}).

Our proposed SCoRL model significantly outperforms the state-of-the-art $S_\text{BERT-base}$ model \cite{PANC} in the test dataset. Specifically, $S_\text{BERT-base}$ achieves a Latency-F1 of only 0.062, whereas our SCoRL, which is designed with a turn-level risk label and speed-control reward, achieved a notably high score of 0.365.
This demonstrates SCoRL’s superior ability to provide accurate and timely detection. 
Moreover, $S_\text{BERT-base}$ shows an imbalance between precision and recall, with a recall of 1.0 but a precision of only 0.047, indicating that it over-identifies conversations as grooming.
In contrast, SCoRL shows a balanced score with a precision of 0.475 and an F1 score of 0.641, highlighting its superior accuracy and precision.
These results demonstrate the effectiveness of SCoRL in distinguishing between grooming and normal conversations.

\paragraph{Result on Downsampled Test Dataset}
To further evaluate performance, we construct a downsampled test dataset by random sampling from normal chats, following the label ratio setup used in previous work \cite{PANC}. Despite the reduced the number of negative chats, SCoRL consistently outperforms the previous state-of-the-art model across all test datasets, demonstrating its robustness and effectiveness.

\paragraph{Impact of Turn-Level Risk Labeling}
As shown in Table \ref{tab: new_metric}, the SFT model, which integrates turn-level risk labels during training, significantly outperforms previous models, achieving an F1-score of 0.211 while maintaining a comparable speed.
This improvement indicates that turn-level risk labeling enables the model to better capture the nuances of dangerous utterances in the conversation.
Interestingly, the SFT (w/o turn-level risk label) model, which assumes the same risk label for all turns in a conversation ($y^{turn}_:=y$), performs worse than both $S_\text{BERT-base}$ and SFT (our) model. This confirms our claim that treating all turns with the same label fails to capture the conversational dynamics needed for accurate grooming detection.

\paragraph{Impact of Speed-Control Reward}
Introducing a speed-control reward mechanism within the reinforcement learning framework improves the model’s timely detection performance. After applying this reward to the SFT model, Latency-F1 increases by over 40\%, reaching 0.365. However, there is a slight reduction in the recall, precision more than doubles, climbing to 0.475, resulting in a twofold increase in the overall F1 score compared to the baseline SFT.
Although there is a minor decrease in detection speed (by 0.062), the significant improvement in accuracy reflects a more optimal balance between timely detection and correct classification. Rather than focusing solely on earlier detection, the model learns to wait until it can confidently differentiate between grooming and normal conversations, resulting in optimal overall turn-level eSPD performance.

\subsection{Discussions}
\paragraph{Chat-Level eSPD}
We further evaluate a traditional eSPD metric that assesses early detection success based solely on the overall chat-level risk label without considering turn-level risk label (Table \ref{tab: previous_metric}). Our experimental results show that both SFT and SCoRL outperform existing models significantly, achieving Latency-F1 scores of 0.489 and 0.401, respectively—around twice the performance of the baselines. 

Interestingly, while the F1 scores of both the baseline $S_\text{BERT-base}$ and our SCoRL model improve under this traditional metric, the degree of improvement differs. The F1 score of $S_\text{BERT-base}$ increases by over four times compared to our metric, while SCoRL shows only a modest 15\% improvement. This suggests that $S_\text{BERT-base}$ often flags early utterances that are not strategically risky, leading to premature detections, as reflected in its higher speed but lower precision. However, the traditional metric mistakenly interprets these premature detections as successful, failing to account for the turn-level context. In contrast, our SCoRL model maintains consistent performance across both turn-level and chat-level metrics, indicating its ability to detect risky utterances at the appropriate time. 

\begin{figure}
\centering
\includegraphics[width=0.9\linewidth]{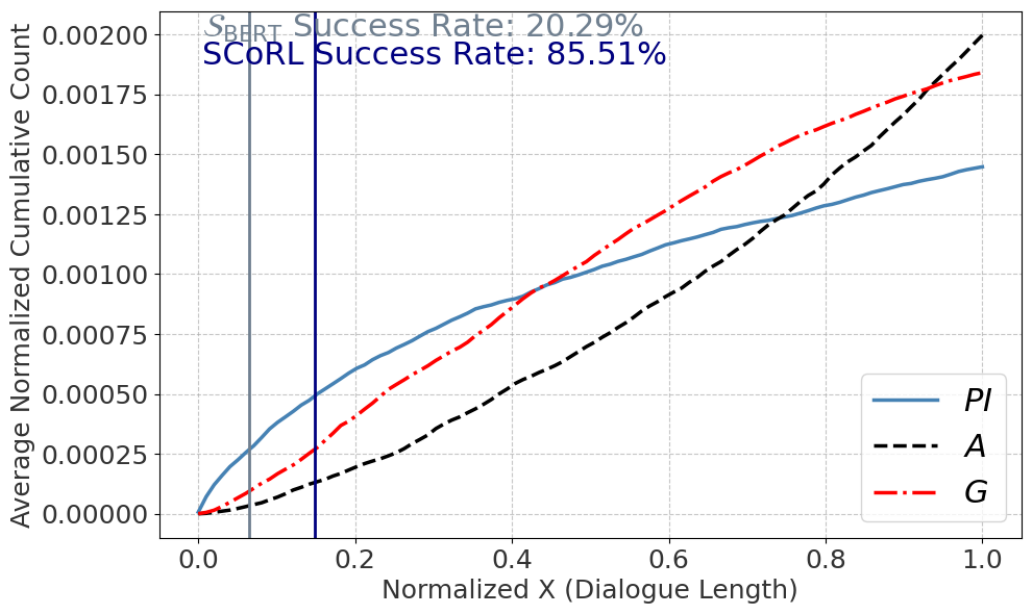}
\caption{Cumulative graph showing the progression of three strategies—\textit{PI}, \textit{A}, and \textit{G}—over time. The x-axis represents the number of turns in the conversation, while the y-axis indicates the cumulative sum of the strategies. The gray vertical line marks the average early detection point of the existing model, and the blue vertical line marks the average detection point achieved by SCoRL.}
\label{fig: visualization1}
\end{figure}

\begin{table}[t]
\scalebox{0.61}{
\begin{tabular}{l|c|cccc}
\toprule
Model & Latency F1 & Precision & Recall & F1 & Speed \\
\midrule
ScoRL (constant reward) & 0.214 & 0.158 & 1.000 & 0.273 & 0.785 \\
ScoRL & \textbf{0.365} & 0.475 & 0.983 & 0.641 & 0.569 \\
\bottomrule
\end{tabular}
}
\caption{Speed control reward ablation study.}\label{tab: ablation_reward}
\end{table}

\paragraph{Earliness Analysis}
To better understand the earliness of our model’s detections, we visualize when SCoRL identifies risky turns during an online grooming conversation, highlighting the shift in LCT turn-level strategy over time.
Figure \ref{fig: visualization1} presents the evolution of three different strategies across an online grooming conversation.
Similar to LCT, we observe that \textit{Exchange of Personal Information} (\textit{PI}) emerge first, followed by \textit{Grooming} (\textit{G}) and \textit{Approach} (\textit{A}) strategies. Both $S_\text{BERT-base}$ and SCoRL achieve early detection within the first 20\% of the conversation on average, demonstrating their capacity for rapid detection. However, while $S_\text{BERT-base}$ detects slightly faster than SCoRL, the accuracy of these early detections differs significantly.
SCoRL achieves an 85.5\% success rate in detecting risky utterances—those associated with \textit{G} or \textit{A} strategies—more than four times higher than $S_\text{BERT-base}$. 
This reveals that $S_\text{BERT-base}$ tends to rush its detections, frequently identifying utterances prematurely. 

\begin{figure}
\centering
\includegraphics[width=\linewidth]{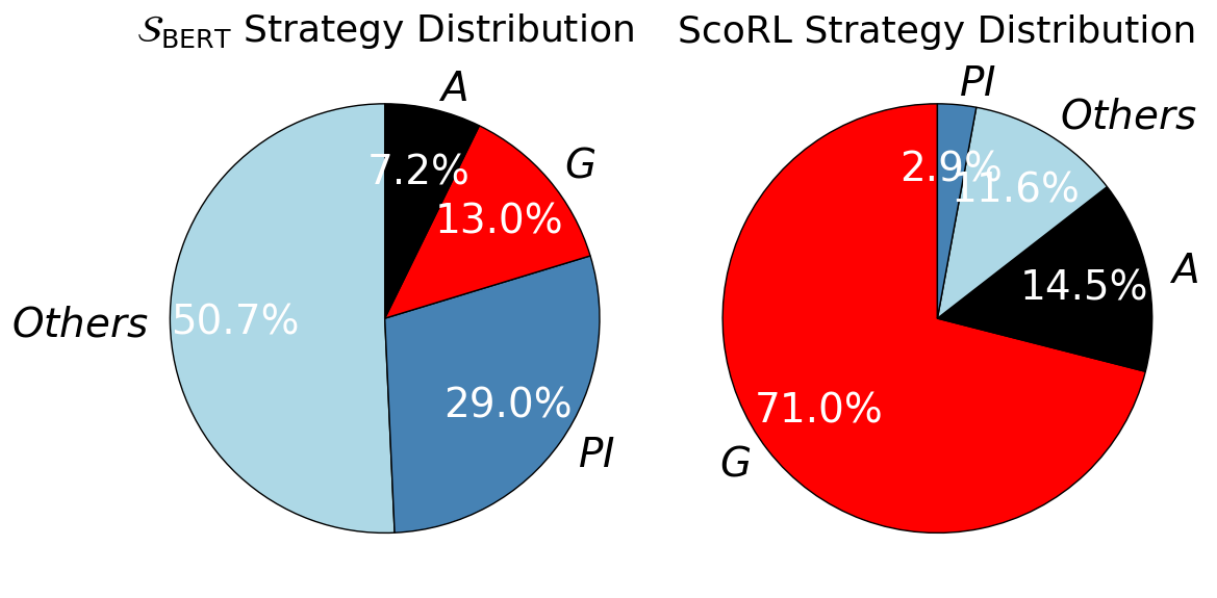}
\caption{Strategy ratio graph for the detected utterance at the point of early detection for each model. The categories include: \textit{PI}, \textit{G}, \textit{A}, and \textit{Others}.}
\label{fig: visualization2}
\end{figure}

When each model triggers early detection, we examine which LCT strategy is associated with the detected turn. Figure \ref{fig: visualization2} presents statistics on the strategies present at the turns where early detection occurs.
Notably, $S_\text{BERT-base}$ performs nearly half of its early detections during utterances labeled as \textit{Others} and 29\% during Exchange of Personal Information (PI), neither of which necessarily indicate turn-level risk. On the other hand, SCoRL detects early in 71\% of cases where the \textit{G} strategy is active, further illustrating its ability to detect danger at the most relevant points in the conversation.

\begin{figure}
\centering
\includegraphics[width=0.9\linewidth]{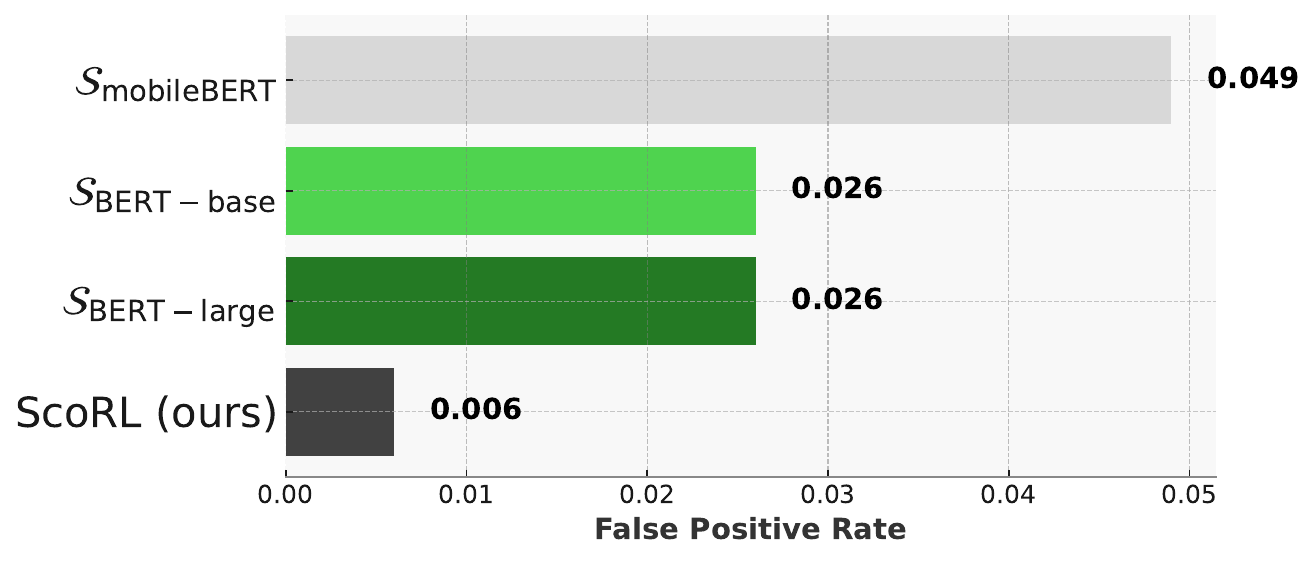}
\caption{False positive rate for previous and our method. This highlights our method’s performance in minimizing false positives rate in low-risk scenarios.}
\label{fig: fpr}
\end{figure}

\paragraph{False Positives Analysis} 
Analyzing false positives in normal conversations is crucial to assessing the model’s ability to avoid mistakenly flagging innocuous interactions. We evaluate the False Positive Rate (FPR), defined as  \[
\text{FPR} = \frac{\text{False Positives}}{\text{False Positives} + \text{True Negatives}},
\] using the PANC dataset for turn-level eSPD (original), as shown in Figure \ref{fig: fpr}. These results underscore the effectiveness of our method in minimizing false positives in normal scenarios.

\paragraph{Speed-Control Reward Analysis} 
To evaluate the effectiveness of our speed control reward mechanism, we compared it against a constant reward approach that assigns $+1$ for successful early detections and $-1$ for failures.
As shown in Table \ref{tab: ablation_reward}, our speed-control reward outperformed the constant reward by a large margin, particularly in precision. As our speed control reward directly optimizes the latency F1, it might converge better than the constant reward.

\begin{figure}
\centering
\includegraphics[width=0.7\linewidth]{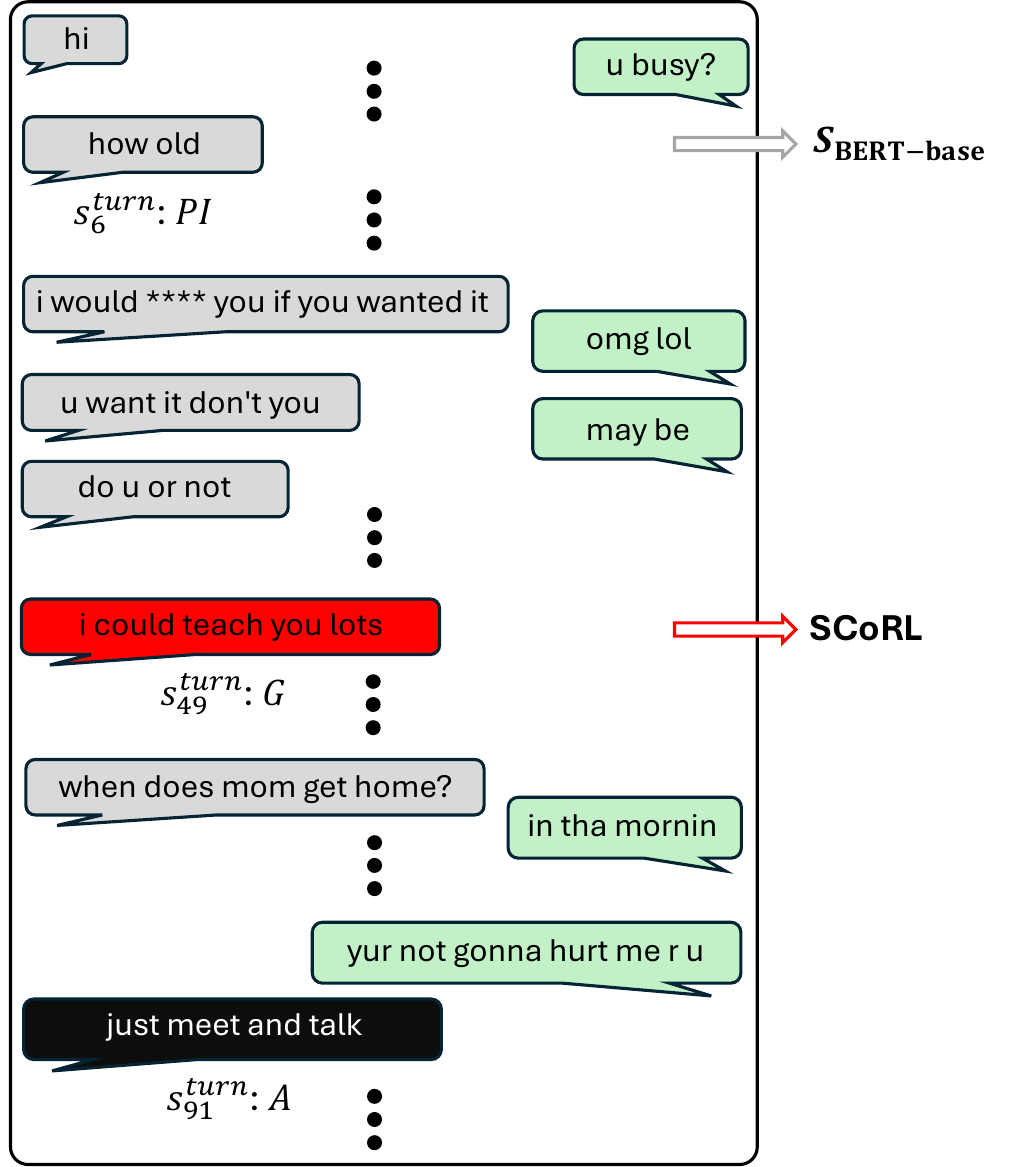}
\caption{Error analysis of early detection of two models in a grooming conversation. The existing model made a false alert too quickly, but SCoRL collects more information and makes early detection sooner. The speech bubble on the left represents the predator, while the one on the right represents the victim. Red indicates the \textit{Grooming} strategy, while black represents the \textit{Approach} strategy.}
\label{fig: error_analysis}
\end{figure}

These results underscore the value of the speed-control reward mechanism. While the constant reward approach encourages quicker detections, it sacrifices accuracy. In contrast, the speed-control reward not only optimizes detection speed but also helps the model identify the optimal time for making accurate distinctions. This balance between speed and precision results in more effective early grooming detection.

\paragraph{eSPD Error Analysis}
Figure \ref{fig: error_analysis} compares the early detection result of $S_\text{BERT-base}$ and SCoRL on the same conversation example data. The conversation follows a typical LCT progression: it starts with \textit{PI}, moves into \textit{G}, and eventually reaches the \textit{A} stage, where the predator reframes sexual activities as a learning experience while attempting to isolate the victim, as described in \cite{mcghee2011learning}. $S_\text{BERT-base}$ triggers early detection prematurely on non-risky utterances. In contrast, SCoRL accurately halts during the \textit{G} stage, leveraging more contextual information, effectively detecting risk before the predator reaches the target. Notably, SCoRL does not simply react to sexual terms; instead, it identifies distinct online grooming strategies that differentiate predatory behavior from general conversations. This ability is further demonstrated in Appendix \ref{sec:FP in normal}, where SCoRL avoids false positives in hard negative normal conversations, even when sexual terms are present.

\section{Conclusion}
In this work, we revisited eSPD by incorporating turn-level risk labels to enhance detection accuracy. We introduced the SCoRL model, designed to identify optimal early detection points using these fine-grained labels, along with a novel evaluation method for turn-level eSPD. Utilizing a dataset annotated with a turn-level risk label based on LCT, we first trained a sub-optimal model using SFT and then improved it by applying our speed-control rewarding mechanism through the REINFORCE algorithm.
Our approach outperformed existing SOTA models, demonstrating the effectiveness of turn-level labels and the speed-control reward. Through extensive experiments, we not only validated the quantitative improvements but also showcased the qualitative advantages of our method.

\section{Limitations}
One limitation of our approach is the variability in the flow of grooming strategy. While online grooming tends to follow similar overarching patterns, the progression of strategies can change fluidly depending on the dynamics between the predator and the victim. However, the current dataset, which includes only 77 conversations with turn-level strategy annotations in the training set, is insufficient to capture the full range of strategic variations. Moreover, ethical concerns arise hiring annotators due to the inherently harmful and toxic nature of the content.
Additionally, our current labeling framework focuses solely on the predator's strategy without considering the victim's state or dialogue acts in response. Incorporating such labels could provide richer context and more valuable insights for improving early sexual predator detection (eSPD).

\section*{Ethics Statement}
Early sexual predator detection (eSPD) is a highly sensitive topic, requiring a careful discussion of the potential implications of such research, the datasets used, and the readiness of eSPD models for real-world applications. The stakes are particularly high for individuals whose conversations are analyzed by eSPD systems. Any deployment of eSPD in live chat systems would involve interaction with vulnerable populations, such as minors, who must be rigorously protected. Both false-negative and false-positive predictions could have serious consequences—either by failing to protect a child or by falsely accusing an innocent chat partner.
Online grooming is illegal in many countries, as are any forms of sexual interaction with minors. In several countries, even obtaining logs of chats involving sexual content with minors is prohibited, making the acquisition or use of real-world data impossible outside of criminal investigations. Nonetheless, online grooming is a real and present issue, underscoring the importance of research aimed at preventing or mitigating it.
For this study, we did not create any new data or conduct experiments involving human subjects. Furthermore, we did not develop or distribute any generative models using ethically sensitive data.

\section*{Acknowledgments}
This work was supported by Institute of Information \& communications Technology Planning \& Evaluation (IITP) grant funded by the Korea government(MSIT) (No.RS-2019-II191906, Artificial Intelligence Graduate School Program(POSTECH)) (5\%)
and by the IITP(Institute of Information \& Coummunications Technology Planning \& Evaluation)-ITRC(Information Technology Research Center) grant funded by the Korea government(Ministry of Science and ICT)(IITP-2025-RS-2024-00437866) (47.5\%).
Additionally, this research was supported by Smart HealthCare for Police Officers Program(www.kipot.or.kr) through the Korea Institutes of Police Technology(KIPoT) funded by the Korean National Police Agency(KNPA, Korea)(No. RS-2022-PT000186) (47.5\%).
\bibliography{custom}

\appendix

\section{Computational Cost in SCoRL}
\label{sec:Computational cost}
During inference, the model processes approximately one turn in 23 \textit{ms} on a single RTX 3090 GPU, utilizing about 1 GB of GPU memory (with a total of 125M parameters). This highlights that the model is computationally efficient and suitable for real-time deployment in real-world systems.
In RL training, the model requires one RTX 3090 GPU and completes training on the full dataset in approximately 5 hours.

\section{False Positive Error in Normal Chats}
\label{sec:FP in normal}
\begin{figure}
\centering
\includegraphics[width=0.7\linewidth]{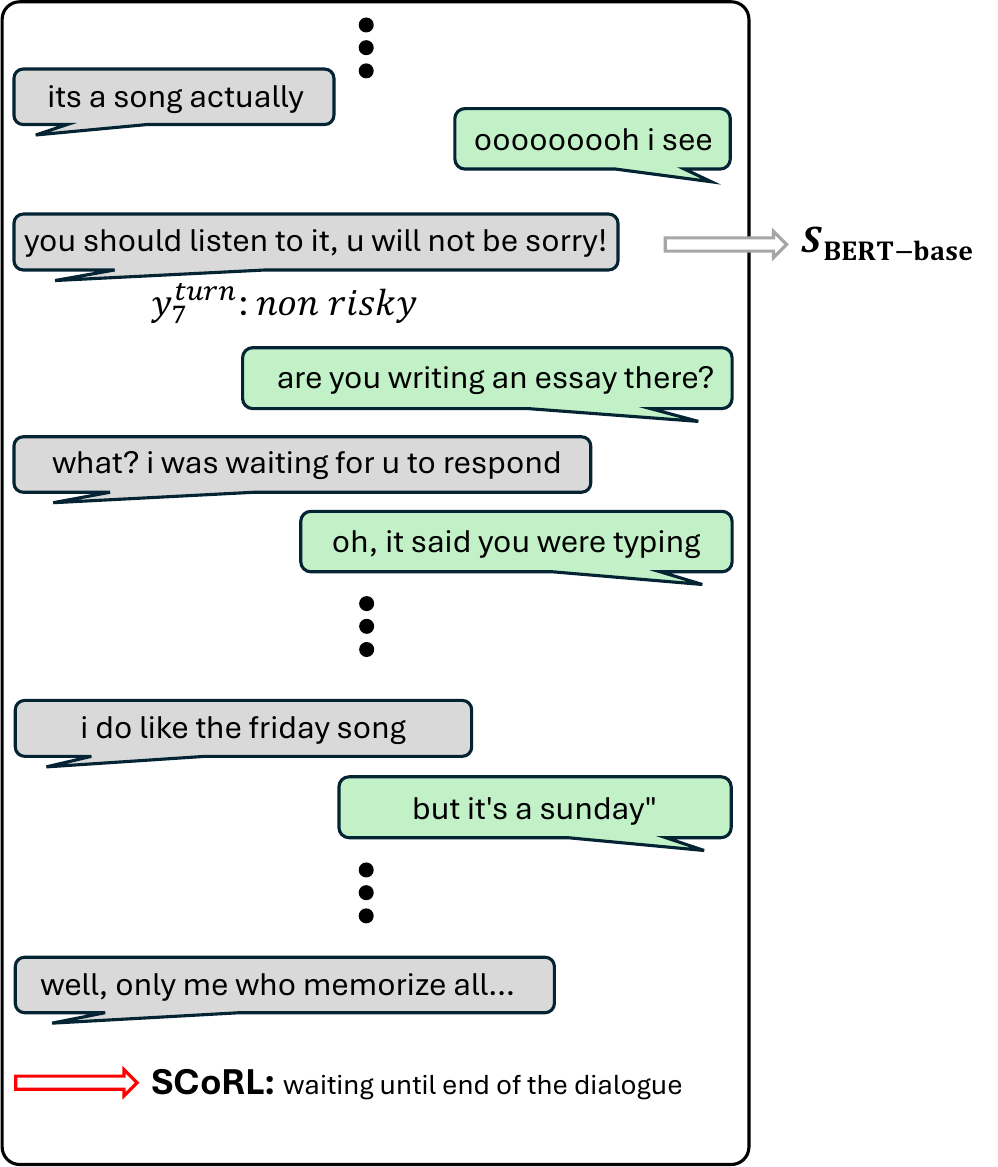}
\caption{This analysis compares early detection performance between two models in a normal conversation.}
\label{fig: neg_chat_FP_normal}
\end{figure}

\begin{figure}
\centering
\includegraphics[width=0.7\linewidth]{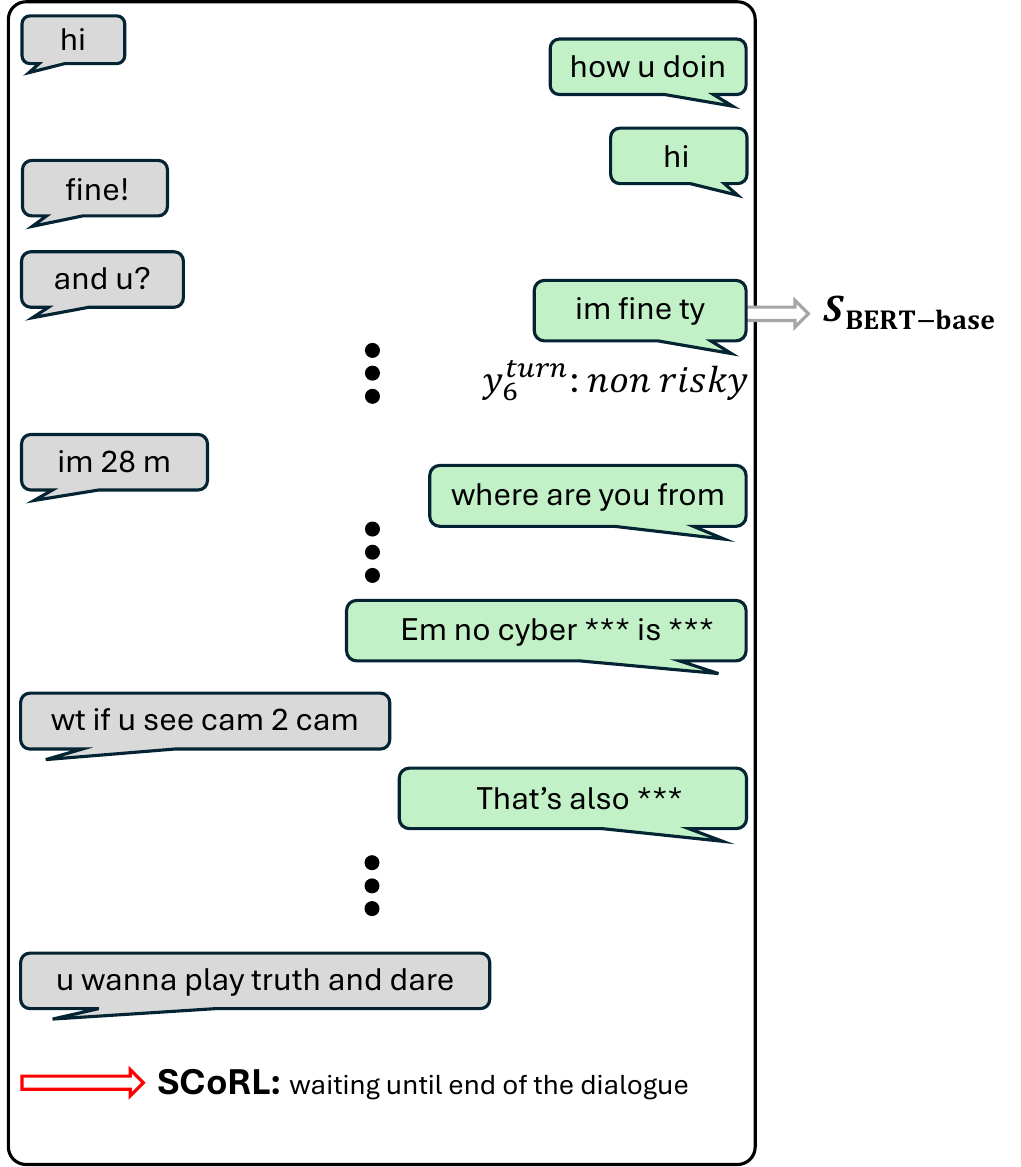}
\caption{This analysis compares the early detection performance of two models in a hard negative normal conversation.}
\label{fig: neg_chat_FP_adult}
\end{figure}

This section examines the false positive errors of previous methods in normal chat settings. In normal conversations, a turn-level eSPD system should not trigger early detection until the conversation ends.
However, Figure \ref{fig: neg_chat_FP_normal} shows that the previous approach incorrectly triggers early detection on non-risky turns, whereas SCoRL correctly waits until the conversation concludes.
Additionally, Figure \ref{fig: neg_chat_FP_adult} presents a hard negative case—a cybersex conversation between consenting adults. Such cases are particularly challenging, as they make it difficult to differentiate online grooming from consensual adult interactions. While the previous model also fails to wait until the end, SCoRL successfully refrains from premature detection, demonstrating its improved ability to handle difficult scenarios.

\end{document}